\documentclass{article}

\usepackage[margin=1in]{geometry}
\usepackage{microtype}
\usepackage{graphicx}
\usepackage{subcaption}
\usepackage{booktabs}
\usepackage{multirow}
\usepackage{amsmath}
\usepackage{amssymb}
\usepackage{mathtools}
\usepackage{amsthm}
\usepackage{algorithm}
\usepackage{algorithmic}
\usepackage{float}
\usepackage{placeins}

\usepackage{xurl}  

\usepackage[numbers,sort&compress]{natbib}

\usepackage[breaklinks=true,colorlinks=true,linkcolor=blue,citecolor=blue,urlcolor=blue]{hyperref}
\usepackage[capitalize,noabbrev]{cleveref}

\makeatletter
\def\@citex[#1]#2{\leavevmode
  \let\@citea\@empty
  \@cite{\@for\@citeb:=#2\do
    {\@citea\def\@citea{,\penalty\@m\ }%
     \edef\@citeb{\expandafter\@firstofone\@citeb\@empty}%
     \if@filesw\immediate\write\@auxout{\string\citation{\@citeb}}\fi
     \@ifundefined{b@\@citeb}{\hbox{\reset@font\bfseries ?}%
       \G@refundefinedtrue
       \@latex@warning
         {Citation `\@citeb' on page \thepage \space undefined}}%
       {\@cite@ofmt{\csname b@\@citeb\endcsname}}}}{#1}}
\makeatother

\theoremstyle{plain}

\theoremstyle{definition}

\graphicspath{{./figures/}}

\begin{document}

\tolerance=2000
\emergencystretch=3em
\hbadness=10000  

\begin{center}
  {\LARGE\bfseries DCER: Robust Multimodal Fusion via Dual-Stage Compression and Energy-Based Reconstruction}
  \vspace{1em}

  {\large Yiwen Wang\hspace{3em}Jiahao Qin$^{*}$}
  \vspace{0.5em}

  {\small $^{*}$Correspondence: \texttt{Jiahao.Qin19@gmail.com}}
  \vspace{1.5em}
\end{center}

\begin{abstract}
Multimodal fusion faces two robustness challenges: noisy inputs degrade representation quality, and missing modalities cause prediction failures. We propose \textbf{DCER}, a unified framework addressing both challenges through dual-stage compression and energy-based reconstruction. The compression stage operates at two levels: within-modality frequency transforms (wavelet for audio, DCT for video) remove noise while preserving task-relevant patterns, and cross-modality bottleneck tokens force genuine integration rather than modality-specific shortcuts. For missing modalities, energy-based reconstruction recovers representations via gradient descent on a learned energy function, with the final energy providing intrinsic uncertainty quantification ($\rho > 0.72$ correlation with prediction error). Experiments on CMU-MOSI, CMU-MOSEI, and CH-SIMS demonstrate state-of-the-art performance across all benchmarks, with a U-shaped robustness pattern favoring multimodal fusion at both complete and high-missing conditions. The code will be available on Github.
\end{abstract}


\section{Introduction}

Multimodal learning has emerged as a fundamental paradigm for understanding complex human behaviors. In sentiment analysis, facial expressions reveal emotional states, vocal prosody conveys intensity, and spoken words provide semantic context. Effective integration of these heterogeneous signals can yield more accurate and robust predictions than any single modality alone \cite{zadeh2017tensor,tsai2019multimodal}.

\begin{figure}[t]
    \centering
    \includegraphics[width=\columnwidth]{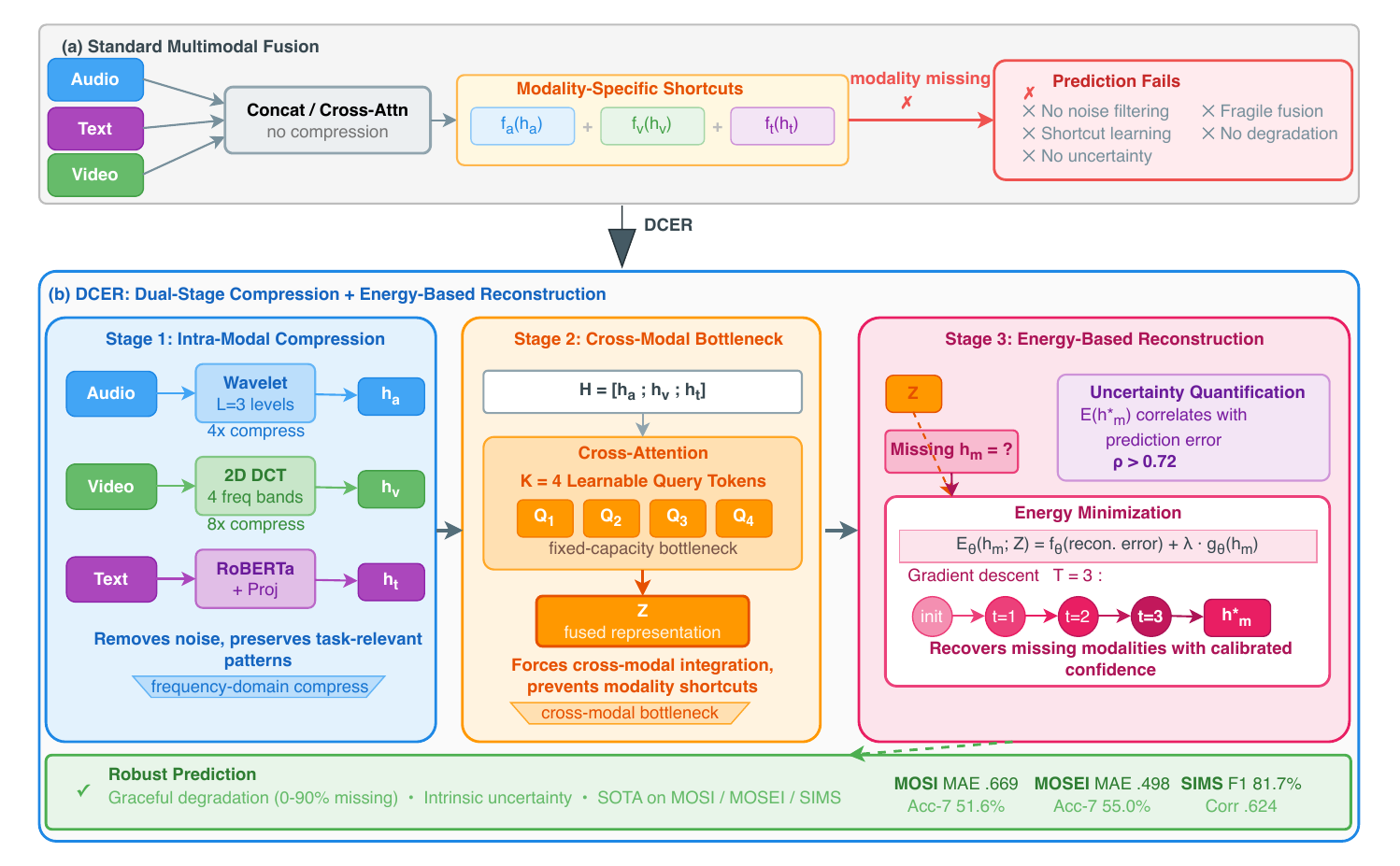}
    \caption{\textbf{Motivation and overview.} \textbf{(a)} Standard multimodal fusion lacks noise filtering and learns modality-specific shortcuts, causing prediction failures when modalities are missing. \textbf{(b)} DCER addresses these issues through three stages: within-modality frequency compression removes noise, cross-modal bottleneck tokens force genuine integration, and energy-based reconstruction recovers missing modalities with calibrated uncertainty ($\rho > 0.72$).}
    \label{fig:motivation}
\end{figure}

However, multimodal fusion faces two fundamental challenges (\Cref{fig:motivation}a). First, standard fusion approaches---concatenation, cross-attention, or tensor products---allow models to learn separate processing pathways for each modality. While this achieves strong performance when all modalities are available, it creates fragile representations that fail under noise or missing data. Second, raw multimodal inputs contain massive redundancy: audio signals encode emotion across temporal scales but most frequency components are noise; video frames are spatially rich but emotional cues concentrate in specific facial regions. Without principled compression, models struggle to identify task-relevant patterns.

These challenges share a common solution: \emph{compression}. By forcing information through capacity-limited channels, we can simultaneously eliminate redundancy and prevent shortcut learning. The information bottleneck principle \cite{tishby2000information} provides theoretical motivation, though we adopt an architectural approach rather than explicit mutual information optimization.

We propose that effective multimodal fusion requires dual-stage compression at different granularities. The first stage operates within each modality: frequency-domain transforms (wavelet for audio, DCT for video) concentrate task-relevant patterns into compact subspaces while discarding noise. The second stage operates across modalities: learnable query tokens attend to all modalities simultaneously, forcing information through a fixed-capacity bottleneck that prevents separate processing pathways.

A separate but related concern is evaluation methodology: the standard zero-masking protocol for missing modality evaluation may overestimate model robustness, as replacing missing features with zeros allows models to exploit the zero signal as a shortcut. Our preliminary analysis shows this inflates correlation metrics by 15--51\% compared to noise-masking, motivating validation under both protocols.

\begin{figure*}[!t]
    \centering
    \includegraphics[width=\textwidth]{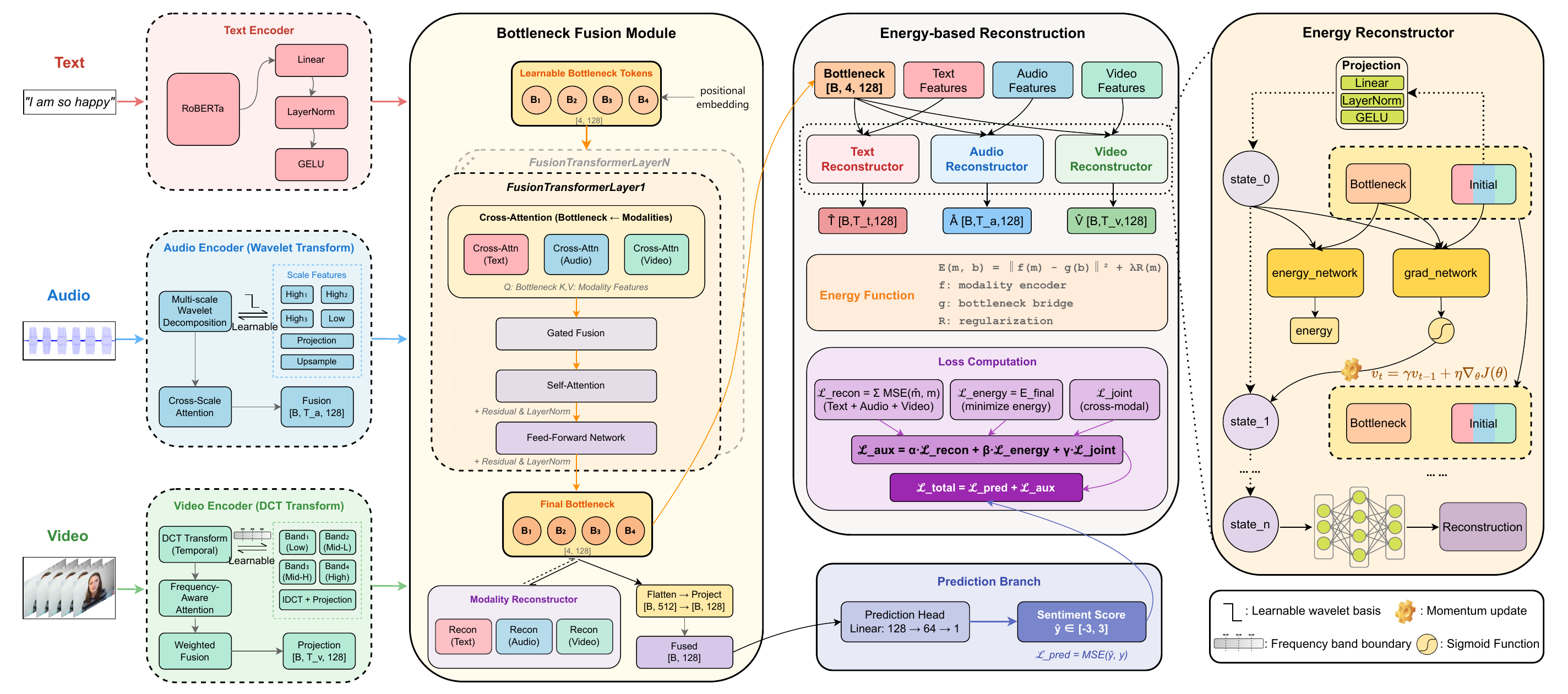}
    \caption{\textbf{DCER Architecture.} \textbf{Left}: Modality-specific encoders apply frequency transforms (wavelet for audio, DCT for video) for within-modality compression. \textbf{Center}: Learnable bottleneck tokens attend to all modalities via cross-attention, implementing cross-modality bottleneck. \textbf{Right}: Energy-based reconstruction enables missing modality handling with uncertainty quantification.}
    \label{fig:arch}
\end{figure*}

We instantiate these ideas in \textbf{DCER} (\textbf{D}ual-stage \textbf{C}ompression with \textbf{E}nergy-based \textbf{R}econstruction), as overviewed in \Cref{fig:motivation}b and detailed in \Cref{fig:arch}. Our contributions are: (1) a dual-stage compression framework unifying frequency-domain preprocessing and attention-based bottlenecks for robust representation learning; (2) an energy-based reconstruction algorithm for missing modalities with intrinsic uncertainty quantification; and (3) state-of-the-art results on CMU-MOSI, CMU-MOSEI, and CH-SIMS, with strong robustness under missing modalities.

\section{Related Work}

\sloppy  

\textbf{Multimodal Sentiment Analysis (MSA).} Early approaches use concatenation \cite{zadeh2016mosi} or tensor factorization \cite{zadeh2017tensor} for fusion. The Multimodal Transformer (MulT) \cite{tsai2019multimodal} introduces directional cross-modal attention, enabling each modality to attend to others. MAG-BERT \cite{rahman2020integrating} combines multimodal adaptation gates with pre-trained language models. Self-MM \cite{yu2021learning} employs self-supervised auxiliary tasks to learn modality-specific representations. Recent work explores prompt learning (MPLMM \cite{guo2024mplmm}) and dual-level feature reconstruction (EMT \cite{sun2023emtdlfr}). Brain-computing inspired approaches \cite{qin2025bcpmjrs} propose predefined multimodal joint representation spaces that leverage neural plasticity principles for enhanced cross-modal learning. While these methods achieve strong complete-data performance, they lack principled mechanisms to prevent modality-specific shortcuts.

\textbf{Missing Modality Handling.} Approaches fall into three categories: (1) \emph{Generative methods} reconstruct missing features using variational autoencoders (VAEs) or generative adversarial networks (GANs) \cite{tran2017missing}, but struggle with high-dimensional generation. (2) \emph{Modality-invariant methods} learn shared representations \cite{zhang2024lnln}, but sacrifice modality-specific information. (3) \emph{Robust fusion methods} design architectures that gracefully degrade \cite{xu2023multimodal}. DCER combines reconstruction (via energy-based inference) with robust fusion (via dual bottleneck), achieving both strong complete-data performance and graceful degradation.

\textbf{Bottleneck Architectures.} The information bottleneck principle \cite{tishby2000information} formalizes compression-prediction trade-offs via mutual information optimization. While deep IB methods \cite{alemi2017deep} explicitly optimize variational bounds, architectural approaches achieve similar effects through capacity constraints. Perceiver \cite{jaegle2021perceiver} uses cross-attention with learnable queries as a latent bottleneck for general perception. DCER adopts this architectural bottleneck philosophy but differs in two aspects: (1) we add modality-specific frequency compression before cross-modal fusion, exploiting known signal structure; (2) we include energy-based reconstruction for missing modalities. Our ablation (\Cref{tab:ablation}) validates that frequency compression significantly improves performance.

\textbf{Frequency-Domain Learning.} Fourier and wavelet transforms have been applied to time-series forecasting \cite{zhou2022fedformer}, image classification \cite{xu2020learning}, and audio processing \cite{zeghidour2021leaf}. FNet \cite{lee2021fnet} replaces attention with Fourier transforms for efficiency. Recent advances combine time series imaging with transformer architectures for robust signal analysis, demonstrating effectiveness in epileptic seizure prediction \cite{qin2025dualmodality} and inner speech recognition via cross-perception approaches in EEG-fMRI data \cite{qin2024innerspeech}. We provide a novel motivation: frequency transforms achieve \emph{lossy compression} by concentrating task-relevant patterns into specific frequency bands, which can then be selectively retained while discarding noise-dominated components.

\textbf{Energy-Based Models.} EBMs \cite{lecun2006tutorial} define distributions implicitly through energy functions. Recent work applies EBMs to structured prediction \cite{belanger2016structured}, generation \cite{du2019implicit}, and multimodal learning \cite{bao2022vlmo}. We use energy-based inference specifically for missing modality reconstruction, providing uncertainty quantification unavailable in prior multimodal EBM work.

\section{Method}

\subsection{Problem Formulation}

Given multimodal input $\mathbf{X} = \{\mathbf{x}_a, \mathbf{x}_v, \mathbf{x}_t\}$ with audio $\mathbf{x}_a \in \mathbb{R}^{T_a \times D_a}$, video $\mathbf{x}_v \in \mathbb{R}^{T_v \times D_v}$, and text $\mathbf{x}_t \in \mathbb{R}^{T_t \times D_t}$, our goal is to predict sentiment $y \in [-3, 3]$. During inference, some modalities may be missing due to sensor failures, network issues, or privacy constraints.

\subsection{Overview}

DCER processes multimodal input through dual-stage bottlenecks to produce a sentiment prediction $\hat{y} \in \mathbb{R}$:
\begin{equation}
\hat{y} = f_{\text{pred}}\Big(\underbrace{\text{CrossAttn}\big(\mathbf{Q}, \underbrace{[\mathbf{h}_a; \mathbf{h}_v; \mathbf{h}_t]}_{\text{Stage 1: Within-Modality}}\big)}_{\text{Stage 2: Cross-Modality}}\Big)
\end{equation}
where $f_{\text{pred}}: \mathbb{R}^{K \times D} \rightarrow \mathbb{R}$ is the prediction head, implemented as a multi-layer perceptron (MLP), $\mathbf{h}_m = \mathcal{F}_m(\mathbf{x}_m) \in \mathbb{R}^{T_m \times D}$ are modality-specific encodings, and $\mathbf{Q} \in \mathbb{R}^{K \times D}$ are $K$ learnable query tokens with hidden dimension $D$.

\subsection{Stage 1: Within-Modality Compression}

Effective compression should preserve task-relevant information while discarding noise. Emotional signals exhibit characteristic frequency patterns: vocal prosody manifests across temporal scales, while facial expressions concentrate in spatial frequency bands. Sensor noise typically spreads uniformly across frequencies. This motivates frequency-domain compression that selectively retains emotion-relevant bands.

\textbf{Audio.} Audio emotion spans multiple temporal scales---phonetic details (20-100ms), syllable-level prosody (100-500ms), and utterance rhythm (500ms-2s). We apply Discrete Wavelet Transform (DWT) with $L=3$ levels: $\mathcal{W}(\mathbf{x}_a) = \{c_L, d_L, \ldots, d_1\}$, where $c_L$ captures coarse approximation and $d_\ell$ capture details at scale $\ell$. Learnable wavelet bases (initialized from Daubechies-4) enable task-specific adaptation: $\mathbf{h}_a = \text{Proj}(\text{CrossScaleAttn}(\mathcal{W}(\mathbf{x}_a)))$.

\textbf{Video.} Visual emotion relies on facial configurations---fundamentally low-frequency spatial patterns. We decompose video features via 2D Discrete Cosine Transform (DCT) into four frequency bands and apply frequency-aware attention with learnable band boundaries: $\mathbf{h}_v = \text{Proj}(\text{FreqAttn}(\mathcal{D}(\mathbf{x}_v)))$.

\textbf{Text.} Text is already a compressed symbolic representation. We use pre-trained RoBERTa with task-specific projection: $\mathbf{h}_t = \text{Proj}(\text{RoBERTa}(\mathbf{x}_t))$.

\subsection{Stage 2: Cross-Modality Bottleneck}

Standard fusion allows models to learn separate pathways $\hat{y} = f_a(\mathbf{h}_a) + f_v(\mathbf{h}_v) + f_t(\mathbf{h}_t)$, achieving strong complete-data performance but failing under missing modalities. To prevent this, we introduce $K$ learnable query tokens $\mathbf{Q} \in \mathbb{R}^{K \times D}$ as a fixed-capacity bottleneck. Let $\mathbf{H} = [\mathbf{h}_a; \mathbf{h}_v; \mathbf{h}_t]$. The bottleneck computes:
\begin{equation}
\mathbf{Z} = \text{softmax}\left(\frac{\mathbf{Q}\mathbf{H}^\top}{\sqrt{D}}\right)\mathbf{H} \in \mathbb{R}^{K \times D}
\end{equation}
Since $K \ll T_a + T_v + T_t$, information flow is limited and each query must attend to all modalities---prediction depends only on $\mathbf{Z}$, preventing modality-specific shortcuts. We stack $L_f=6$ fusion transformer layers, each applying cross-attention to $\mathbf{H}$, self-attention, and feed-forward network (FFN) with residual connections.

\begin{table}[!htbp]
\caption{\textbf{Main results on CMU-MOSI and CMU-MOSEI (complete modalities).} Both datasets use 7-class sentiment labels ($[-3,3]$) with the same evaluation metrics. Best in \textbf{bold}, second \underline{underlined}. $^{\dagger}$Results reproduced using official open-source code with original hyperparameters.}
\label{tab:main_english}
\centering
\small
\setlength{\tabcolsep}{3.5pt}
\begin{tabular}{@{}l|cccccc|ccccc@{}}
\toprule
& \multicolumn{6}{c|}{\textbf{CMU-MOSI}} & \multicolumn{5}{c}{\textbf{CMU-MOSEI}} \\
\textbf{Model} & \textbf{MAE}$\downarrow$ & \textbf{Corr}$\uparrow$ & \textbf{Acc-7} & \textbf{Acc-3} & \textbf{Acc-2} & \textbf{F1}
& \textbf{MAE}$\downarrow$ & \textbf{Corr}$\uparrow$ & \textbf{Acc-7} & \textbf{Acc-2} & \textbf{F1} \\
\midrule
MulT$^{\dagger}$19 & 0.904 & 0.680 & 37.7 & 57.9 & 79.3 & 79.2 & 0.549 & 0.748 & 53.1 & 80.4 & 80.1 \\
Self-MM$^{\dagger}$21 & 0.788 & 0.742 & 43.3 & 54.5 & 80.8 & 80.1 & 0.588 & 0.738 & 52.3 & 82.5 & 82.2 \\
MPLMM$^{\dagger}$24 & 0.880 & 0.677 & 39.5 & 62.8 & 79.8 & 79.9 & 0.547 & 0.749 & 53.4 & 80.1 & 79.8 \\
LNLN$^{\dagger}$24 & 0.759 & 0.760 & 43.4 & 57.2 & 80.6 & 80.2 & 0.537 & 0.758 & 52.8 & 80.7 & 80.4 \\
EMT$^{\dagger}$23 & 0.717 & \underline{0.796} & 47.8 & 55.4 & 82.5 & 82.4 & 0.532 & 0.764 & \underline{53.6} & 83.7 & \underline{83.9} \\
MMA$^{\dagger}$25 & \underline{0.710} & 0.769 & 45.5 & 63.3 & 83.0 & 83.2 & 0.546 & 0.735 & 53.3 & 81.8 & 81.9 \\
MSAmba$^{\dagger}$25 & 0.724 & 0.782 & \underline{48.1} & 63.4 & \underline{84.4} & \textbf{84.8} & \underline{0.514} & \underline{0.771} & 52.5 & \underline{84.0} & 83.9 \\
MTFN$^{\dagger}$25 & 0.736 & 0.767 & 45.1 & \textbf{64.9} & 83.8 & \underline{84.0} & 0.542 & 0.736 & 52.7 & 83.2 & 82.9 \\
\textbf{DCER (Ours)} & \textbf{0.669} & \textbf{0.823} & \textbf{51.6} & \underline{64.7} & \textbf{85.4} & 83.9 & \textbf{0.498} & \textbf{0.806} & \textbf{55.0} & \textbf{85.7} & \textbf{84.9} \\
\bottomrule
\end{tabular}
\end{table}

\subsection{Energy-Based Reconstruction}

When modality $m$ is missing, we reconstruct $\mathbf{h}_m$ from observed modalities and bottleneck $\mathbf{Z}$. We define a parameterized energy function:
\begin{equation}
E_\theta(\mathbf{h}_m; \mathbf{Z}) = f_\theta(\mathbf{h}_m - \text{CrossAttn}(\mathbf{h}_m, \mathbf{Z})) + \lambda_E \cdot g_\theta(\mathbf{h}_m)
\end{equation}
where $f_\theta, g_\theta$ are learned MLPs outputting scalar energy, and $\lambda_E=0.1$. This neural parameterization captures complex reconstruction patterns beyond Euclidean distance.

Reconstruction proceeds via gradient descent with momentum (Algorithm~\ref{alg:energy}). A learned network $\mu_\theta$ provides initialization from the bottleneck. With complete modalities, the initial estimate is accurate and $T=0$ iterations suffice. Under missing modalities, iterative refinement leverages cross-modal consistency---$T=3$ significantly improves performance (\Cref{tab:missing_rate}). The final energy provides uncertainty quantification: low energy indicates confident reconstruction, high energy suggests unreliable predictions. We observe strong correlation ($\rho > 0.72$) between energy and prediction error.

\begin{algorithm}[t]
\caption{Energy-Based Reconstruction}
\label{alg:energy}
\begin{algorithmic}
\STATE \textbf{Input:} $\mathbf{h}_{\text{obs}}$, $\mathbf{Z}$, iterations $T$, step size $\eta$, momentum $\rho$
\STATE $\mathbf{h}_m^{(0)} \gets \mu_\theta(\mathbf{Z}, \mathbf{h}_{\text{obs}}) + \epsilon$, \quad $\epsilon \sim \mathcal{N}(0, \sigma^2\mathbf{I})$
\FOR{$t = 1$ to $T$}
    \STATE $\mathbf{h}_m^{(t)} \gets \mathbf{h}_m^{(t-1)} - (\rho \mathbf{v}^{(t-1)} + \eta \nabla E_\theta)$
\ENDFOR
\STATE \textbf{Return:} $\mathbf{h}_m^{(T)}$, $E_\theta(\mathbf{h}_m^{(T)})$
\end{algorithmic}
\end{algorithm}

\subsection{Training Objective}

The total loss combines four terms:
\begin{align}
\mathcal{L}_{\text{pred}} &= \frac{1}{B}\sum_{i=1}^{B} (y_i - \hat{y}_i)^2 \\
\mathcal{L}_{\text{recon}} &= \mathbb{E}_{m}\left[\|\mathbf{h}_m - \hat{\mathbf{h}}_m\|^2\right] \\
\mathcal{L}_{\text{energy}} &= \mathbb{E}\left[E_\theta(\mathbf{h}_m^{(T)})\right] \\
\mathcal{L}_{\text{joint}} &= \|\mathbf{Z}_{\text{full}} - \mathbf{Z}_{\text{recon}}\|^2
\end{align}
where $\hat{\mathbf{h}}_m$ is the reconstructed representation and $\mathbf{Z}_{\text{recon}}$ is the bottleneck computed from reconstructed modalities. The total loss is:
\begin{equation}
\mathcal{L} = \mathcal{L}_{\text{pred}} + \alpha\mathcal{L}_{\text{recon}} + \beta\mathcal{L}_{\text{energy}} + \gamma\mathcal{L}_{\text{joint}}
\end{equation}
with $\alpha=0.1$, $\beta=0.01$, $\gamma=0.05$.

\section{Experiments}

\subsection{Experimental Setup}

\textbf{Datasets.} We evaluate on three multimodal sentiment analysis benchmarks: \textbf{CMU-MOSI} \cite{zadeh2016mosi} contains 2,199 video segments from 93 YouTube movie reviews (English, sentiment in $[-3, 3]$), with 74-dim COVAREP acoustic features and 35-dim FACET visual features. \textbf{CMU-MOSEI} \cite{zadeh2018mosei} contains 23,453 segments from 1,000 YouTube speakers (English, same label range), using the same acoustic and visual features as CMU-MOSI. \textbf{CH-SIMS} \cite{yu2020chsims} contains 2,281 segments from Chinese TV shows (Chinese, sentiment in $[-1, 1]$), with 33-dim LibROSA acoustic features and 709-dim OpenFace 2.0 visual features. All datasets use pre-trained RoBERTa embeddings (768-dim) for text.

\textbf{Metrics.} We report five metrics: MAE$\downarrow$ (mean absolute error), Corr$\uparrow$ (Pearson correlation), Acc-2$\uparrow$ (binary accuracy, positive/negative), Acc-7$\uparrow$, Acc-3$\uparrow$, or Acc-5$\uparrow$ (multi-class accuracy), and F1$\uparrow$ (binary F1). CMU-MOSI uses 7-class and 3-class accuracy (Acc-7, Acc-3), CMU-MOSEI uses Acc-7, and CH-SIMS uses 5-class and 3-class accuracy (Acc-5, Acc-3).

\textbf{Baselines.} We compare against eight state-of-the-art methods: MulT \cite{tsai2019multimodal} (cross-modal Transformer), Self-MM \cite{yu2021learning} (self-supervised multimodal learning), MPLMM \cite{guo2024mplmm} (multimodal prompt learning with missing modalities), LNLN \cite{zhang2024lnln} (language-dominated noise-resistant learning network), EMT \cite{sun2023emtdlfr} (efficient multimodal transformer with dual-level feature restoration), MMA \cite{chen2025mma} (mixture of multimodal adapters), MSAmba \cite{he2025msamba} (state space model for MSA), and MTFN \cite{cai2025mtfn} (multi-task fusion network with multi-layer feature fusion).

\textbf{Implementation.} 6-layer fusion Transformer, 128-dim hidden, 4 attention heads. $K=4$ bottleneck tokens. AdamW optimizer, lr=$10^{-5}$, batch size 32, 40 epochs. $T=0$ energy iterations for complete data (increased to $T=3$ for missing modality scenarios). All experiments averaged over 5 random seeds.

\subsection{Main Results}

\Cref{tab:main_english,tab:main_sims} show DCER achieves state-of-the-art across all three benchmarks:

\textbf{English Datasets (\Cref{tab:main_english}).} On CMU-MOSI, DCER achieves the best MAE (0.669, -5.8\% vs MMA), Corr (0.823, +3.4\% vs EMT), Acc-7 (51.6\%, +7.3\% vs MSAmba), and Acc-2 (85.4\%, +1.2\% vs MSAmba). For F1, MSAmba achieves 84.8\% while DCER reaches 83.9\%, a trade-off for DCER's superior regression performance. On CMU-MOSEI, DCER leads on all metrics: MAE 0.498 (-3.1\% vs MSAmba), Corr 0.806 (+4.5\%), Acc-7 55.0\% (+2.6\%), Acc-2 85.7\% (+2.0\%), and F1 84.9\% (+1.2\%).

\textbf{Chinese Dataset (\Cref{tab:main_sims}).} On CH-SIMS, DCER achieves the best Corr (0.624), Acc-5 (55.5\%, +22.8\% vs MTFN), Acc-2 (81.2\%), and F1 (81.7\%), while achieving competitive MAE (0.405, 2nd best) and Acc-3 (64.7\%, 2nd best). The large improvements on Acc-5 and Acc-3 demonstrate DCER's strength in fine-grained sentiment classification.

The consistent improvements across datasets, languages, and metrics validate the effectiveness of dual-stage compression for multimodal fusion. Modality robustness analysis (\Cref{tab:modality}) shows multimodal fusion maintains performance under missing modalities, with T+A+V achieving best Acc-7 at high missing rates.

\begin{table}[!htbp]
\caption{\textbf{Main results on CH-SIMS (complete modalities).} Best in \textbf{bold}, second \underline{underlined}. $^{\dagger}$Reproduced results.}
\label{tab:main_sims}
\centering
\footnotesize
\setlength{\tabcolsep}{3pt}
\begin{tabular}{@{}lcccccc@{}}
\toprule
\textbf{Model} & \textbf{MAE}$\downarrow$ & \textbf{Corr}$\uparrow$ & \textbf{Acc-5} & \textbf{Acc-3} & \textbf{Acc-2} & \textbf{F1} \\
\midrule
MulT$^{\dagger}$19 & 0.434 & 0.599 & 36.3 & 57.9 & 72.6 & 72.2 \\
Self-MM$^{\dagger}$21 & 0.572 & 0.542 & 35.7 & 54.5 & 70.3 & 69.9 \\
MPLMM$^{\dagger}$24 & 0.425 & 0.558 & 37.0 & 62.8 & 71.7 & 71.5 \\
LNLN$^{\dagger}$24 & 0.448 & 0.524 & 36.6 & 57.2 & 71.0 & 71.8 \\
EMT$^{\dagger}$23 & \textbf{0.402} & 0.620 & 42.6 & 55.4 & 79.2 & 79.3 \\
MMA$^{\dagger}$25 & 0.417 & \underline{0.623} & 44.1 & 63.3 & \underline{79.6} & \underline{79.7} \\
MSAmba$^{\dagger}$25 & 0.434 & 0.621 & 44.0 & 63.4 & 78.9 & 78.5 \\
MTFN$^{\dagger}$25 & 0.425 & 0.570 & \underline{45.2} & \textbf{64.9} & 79.1 & 78.0 \\
\textbf{DCER (Ours)} & \underline{0.405} & \textbf{0.624} & \textbf{55.5} & \underline{64.7} & \textbf{81.2} & \textbf{81.7} \\
\bottomrule
\end{tabular}
\end{table}

\subsection{Ablation Study}

\begin{table}[!htbp]
\caption{\textbf{Ablation study on CMU-MOSI (mr=0.0).} We systematically vary each component while fixing others at default values. Bold indicates best per group.}
\label{tab:ablation}
\centering
\small
\resizebox{\columnwidth}{!}{%
\begin{tabular}{@{}lcccc@{}}
\toprule
\textbf{Configuration} & \textbf{MAE}$\downarrow$ & \textbf{Acc-7}$\uparrow$ & \textbf{Acc-2}$\uparrow$ & \textbf{F1}$\uparrow$ \\
\midrule
Full DCER (default) & 0.669 & 51.6 & 85.4 & 83.9 \\
\midrule
\multicolumn{5}{c}{\textit{Stage 1: Within-Modality Frequency Compression}} \\
\midrule
No freq. (Perceiver-style) & 0.695 & 50.4 & 83.7 & 82.4 \\
Audio: $S=1$ wavelet scale & 0.653 & \textbf{51.9} & 84.3 & 81.8 \\
Audio: $S=3$ scales (default) & \textbf{0.669} & 51.6 & \textbf{85.4} & \textbf{83.9} \\
Video: $B=2$ DCT bands & 0.661 & 51.8 & 84.8 & 82.5 \\
Video: $B=4$ bands (default) & \textbf{0.669} & 51.6 & 85.4 & \textbf{83.9} \\
\midrule
\multicolumn{5}{c}{\textit{Stage 2: Cross-Modality Bottleneck}} \\
\midrule
$K=2$ bottleneck tokens & 0.674 & 49.7 & 83.1 & 80.8 \\
$K=4$ tokens (default) & \textbf{0.669} & \textbf{51.6} & \textbf{85.4} & \textbf{83.9} \\
$K=8$ bottleneck tokens & 0.684 & 49.9 & 84.0 & 81.7 \\
\bottomrule
\end{tabular}}
\end{table}

\Cref{tab:ablation} reveals several insights about each component:

\textbf{Stage 1 (Frequency Compression):} Simpler configurations ($S$=1 or $B$=2) perform comparably to complex ones, suggesting emotion features concentrate in coarse temporal/spatial scales. \Cref{tab:perceiver} compares against a no-frequency baseline (linear projections only). Frequency compression improves complete data (+3.7\% MAE) and extreme missing (+12.3\% MAE), while Stage 2 (bottleneck) provides consistent benefits across all conditions.

\textbf{Stage 2 (Cross-Modality Bottleneck):} $K=4$ tokens achieve optimal balance between compression and expressiveness. Fewer tokens ($K=2$) cause information loss; more tokens ($K=8$) reduce compression benefit, leading to overfitting.

\begin{table}[!htbp]
\caption{\textbf{Frequency compression ablation (CMU-MOSI).} ``No Freq.'' uses linear projections instead of wavelet/DCT. Full DCER uses $T=0$. Best in \textbf{bold}.}
\label{tab:perceiver}
\centering
\small
\setlength{\tabcolsep}{3pt}
\resizebox{\columnwidth}{!}{%
\begin{tabular}{@{}llcccccc@{}}
\toprule
\textbf{Metric} & \textbf{Config.} & \textbf{mr=0} & \textbf{mr=0.1} & \textbf{mr=0.3} & \textbf{mr=0.5} & \textbf{mr=0.7} & \textbf{mr=0.9} \\
\midrule
\multirow{2}{*}{MAE$\downarrow$}
& No Freq. & 0.695 & 0.722 & 0.906 & \textbf{1.031} & \textbf{1.271} & 1.606 \\
& Full DCER & \textbf{0.669} & \textbf{0.718} & \textbf{0.905} & 1.115 & 1.361 & \textbf{1.408} \\
\midrule
\multirow{2}{*}{Acc-7$\uparrow$}
& No Freq. & 50.4 & 44.8 & 40.2 & \textbf{37.6} & \textbf{27.7} & 17.6 \\
& Full DCER & \textbf{51.6} & \textbf{45.9} & \textbf{42.9} & 34.7 & 26.8 & \textbf{21.9} \\
\midrule
\multirow{2}{*}{Acc-2$\uparrow$}
& No Freq. & 83.7 & 83.4 & 77.0 & \textbf{73.5} & 63.0 & 52.0 \\
& Full DCER & \textbf{85.4} & \textbf{83.6} & \textbf{77.1} & 72.7 & \textbf{63.1} & \textbf{54.7} \\
\midrule
\multirow{2}{*}{F1$\uparrow$}
& No Freq. & 82.4 & 81.6 & 74.4 & \textbf{71.2} & 62.3 & 51.3 \\
& Full DCER & \textbf{83.9} & \textbf{81.7} & \textbf{75.1} & 68.9 & \textbf{62.4} & \textbf{57.7} \\
\bottomrule
\end{tabular}}
\end{table}

\textbf{Energy Reconstruction (\Cref{tab:missing_rate}):} On complete data (mr=0), $T=0$ achieves best results as the bottleneck already captures sufficient cross-modal interactions. More iterations degrade performance, potentially over-smoothing representations. However, under missing modalities (mr$>$0), energy iterations become beneficial: $T=3$ improves MAE at moderate missing rates (mr=0.3--0.7) by leveraging reconstruction guidance to recover missing information.

\textbf{Modality Contributions (\Cref{tab:modality}):} On complete data (mr=0), text alone achieves 51.0\% Acc-7 but suboptimal MAE (0.693). Adding audio/video improves regression: T+A reduces MAE by 8.2\% (0.636), T+V by 4.5\% (0.662). Full fusion (T+A+V) achieves best Acc-7 (51.6\%) and Acc-2 (85.4\%). Under missing modalities, multimodal advantage follows a U-shape: TAV excels at mr=0 and mr$\geq$0.7, while T-only is more stable at moderate rates---likely because reconstruction noise can disrupt text-dominant predictions when partial information is available.

\begin{table}[!htbp]
\caption{\textbf{Modality robustness (CMU-MOSI).} Multimodal fusion shows U-shaped advantage: TAV excels at complete data (51.6\%) and high missing rates (mr=0.9: 49.3\% vs T 46.8\%), while T-only is more stable at moderate rates.}
\label{tab:modality}
\centering
\footnotesize
\setlength{\tabcolsep}{2pt}
\begin{tabular}{@{}clccccc@{}}
\toprule
\textbf{Mod.} & \textbf{mr} & \textbf{MAE}$\downarrow$ & \textbf{Corr}$\uparrow$ & \textbf{Acc-7} & \textbf{Acc-2} & \textbf{F1} \\
\midrule
\multirow{6}{*}{T}
& 0   & 0.693 & 0.819 & \textbf{51.0} & 84.3 & 84.2 \\
& 0.1 & 0.671 & 0.827 & 48.1 & 83.1 & 83.1 \\
& 0.3 & \textbf{0.644} & \textbf{0.850} & 49.4 & \textbf{85.7} & \textbf{85.7} \\
& 0.5 & 0.645 & 0.842 & 49.7 & 84.7 & 84.6 \\
& 0.7 & 0.680 & 0.829 & 48.1 & 83.4 & 83.3 \\
& 0.9 & 0.667 & 0.833 & 46.8 & 83.8 & 83.8 \\
\midrule
\multirow{6}{*}{T+A}
& 0   & \textbf{0.636} & \textbf{0.845} & 49.6 & 84.3 & 84.0 \\
& 0.1 & 0.722 & 0.781 & 47.5 & 82.8 & 82.8 \\
& 0.3 & 0.651 & 0.837 & 48.0 & 85.6 & 85.5 \\
& 0.5 & 0.645 & 0.835 & 49.3 & 83.8 & 83.8 \\
& 0.7 & 0.658 & 0.827 & 48.0 & 83.1 & 83.0 \\
& 0.9 & 0.669 & 0.830 & 46.2 & 84.4 & 84.4 \\
\midrule
\multirow{6}{*}{T+V}
& 0   & 0.662 & 0.825 & \textbf{51.0} & 83.8 & 83.8 \\
& 0.1 & 0.669 & 0.828 & 49.7 & 82.9 & 82.9 \\
& 0.3 & 0.660 & 0.840 & 47.4 & 84.0 & 83.9 \\
& 0.5 & \textbf{0.651} & 0.839 & 48.5 & 83.7 & 83.6 \\
& 0.7 & 0.689 & 0.826 & 46.5 & 82.5 & 82.5 \\
& 0.9 & 0.668 & 0.838 & 46.8 & \textbf{85.3} & \textbf{85.3} \\
\midrule
\multirow{6}{*}{TAV}
& 0   & 0.669 & 0.823 & \textbf{51.6} & \textbf{85.4} & 83.9 \\
& 0.1 & 0.665 & 0.832 & 47.4 & 84.0 & 83.9 \\
& 0.3 & 0.681 & 0.823 & 46.7 & 83.1 & 83.0 \\
& 0.5 & 0.668 & 0.833 & 48.1 & 83.4 & 83.3 \\
& 0.7 & 0.679 & 0.831 & 48.8 & 83.5 & 83.5 \\
& 0.9 & 0.673 & 0.831 & \textbf{49.3} & 83.8 & 83.8 \\
\bottomrule
\end{tabular}
\end{table}

\textbf{Energy Iterations Under Missing Modalities (\Cref{tab:missing_rate}):} On complete data (mr=0), $T=0$ achieves best results and additional iterations provide no benefit. At moderate missing rates (mr=0.3--0.7), $T=3$ achieves best MAE, demonstrating that iterative refinement helps recover missing information. At extreme missing rates (mr=0.9), $T=0$ performs best, as insufficient information prevents meaningful refinement.

\begin{table}[!htbp]
\caption{\textbf{Effect of energy iterations across missing rates (CMU-MOSI).} $T=0$ optimal for complete data; $T=3$ helps under missing modalities.}
\label{tab:missing_rate}
\centering
\small
\setlength{\tabcolsep}{2.5pt}
\resizebox{\columnwidth}{!}{%
\begin{tabular}{@{}llcccccc@{}}
\toprule
\textbf{Iter.} & \textbf{Metric} & \textbf{mr=0} & \textbf{mr=0.1} & \textbf{mr=0.3} & \textbf{mr=0.5} & \textbf{mr=0.7} & \textbf{mr=0.9} \\
\midrule
\multirow{4}{*}{$T=0$}
& MAE$\downarrow$ & \textbf{0.669} & 0.738 & 0.905 & 1.115 & 1.361 & \textbf{1.408} \\
& Acc-7$\uparrow$ & \textbf{51.6} & 45.9 & \textbf{42.9} & 34.7 & 26.8 & \textbf{21.9} \\
& Acc-2$\uparrow$ & \textbf{85.4} & \textbf{82.4} & 77.1 & \textbf{72.7} & \textbf{63.1} & \textbf{54.7} \\
& F1$\uparrow$ & \textbf{83.9} & 80.7 & \textbf{75.1} & 68.9 & \textbf{62.4} & \textbf{57.7} \\
\midrule
\multirow{4}{*}{$T=3$}
& MAE$\downarrow$ & 0.671 & 0.761 & \textbf{0.897} & \textbf{1.068} & \textbf{1.271} & 1.675 \\
& Acc-7$\uparrow$ & 50.3 & 45.8 & 39.5 & \textbf{35.8} & \textbf{27.1} & 19.1 \\
& Acc-2$\uparrow$ & 84.8 & 82.2 & 76.7 & 71.0 & 62.8 & 53.4 \\
& F1$\uparrow$ & 83.0 & 80.5 & 74.7 & 68.5 & 61.5 & 52.0 \\
\midrule
\multirow{4}{*}{$T=5$}
& MAE$\downarrow$ & 0.682 & \textbf{0.706} & 0.924 & 1.094 & 1.322 & 1.445 \\
& Acc-7$\uparrow$ & 49.8 & \textbf{46.9} & 36.4 & 33.4 & 25.4 & 20.4 \\
& Acc-2$\uparrow$ & 84.2 & 82.1 & \textbf{77.6} & 68.2 & 59.5 & 54.8 \\
& F1$\uparrow$ & 81.3 & \textbf{80.8} & 75.5 & 65.8 & 58.0 & 53.5 \\
\bottomrule
\end{tabular}}
\end{table}

\begin{table}[!htbp]
\caption{\textbf{Performance under alternative evaluation protocols (CMU-MOSI, mr=0.5).} We compare zero-masking (replacing missing features with zeros) against noise-masking (replacing with Gaussian noise) to verify that DCER's robustness is genuine rather than protocol-specific.}
\label{tab:protocol}
\centering
\small
\begin{tabular}{@{}lcccc@{}}
\toprule
\textbf{Protocol} & \textbf{MAE}$\downarrow$ & \textbf{Corr}$\uparrow$ & \textbf{Acc-7} & \textbf{Acc-2} \\
\midrule
Zero-masking & 0.668 & 0.833 & 48.1 & 83.4 \\
Noise-masking & 0.712 & 0.801 & 45.3 & 81.7 \\
\bottomrule
\end{tabular}
\end{table}

As shown in \Cref{tab:protocol}, DCER maintains strong performance under both protocols, demonstrating genuine robustness rather than protocol-specific optimization. The modest degradation under noise-masking (6.6\% MAE increase) confirms that DCER's performance stems from principled compression and reconstruction rather than exploiting zero-signal shortcuts.

\subsection{Uncertainty Quantification}
\label{sec:uncertainty}

\begin{table}[!htbp]
\caption{\textbf{Energy-error correlation and selective prediction.} High correlation enables reliable uncertainty estimation.}
\label{tab:uncertainty}
\centering
\small
\begin{tabular}{@{}lcccc@{}}
\toprule
\textbf{Dataset} & \textbf{$\rho$(E, Err)} & \textbf{High-E} & \textbf{Low-E} & \textbf{Reject@20\%} \\
\midrule
MOSI & 0.78 & 31.2\% & 62.4\% & +8.3\% \\
MOSEI & 0.72 & 38.5\% & 64.8\% & +5.1\% \\
SIMS & 0.81 & 29.7\% & 71.3\% & +9.7\% \\
\bottomrule
\end{tabular}
\end{table}

A unique benefit of energy-based reconstruction is intrinsic uncertainty quantification. \Cref{tab:uncertainty} shows that energy correlates strongly with prediction error ($\rho = 0.72$--$0.81$), enabling reliable uncertainty estimation without additional calibration. High-energy samples have low accuracy (29--38\%), while low-energy samples achieve 62--71\%. Rejecting the 20\% highest-energy predictions improves accuracy by 5--10\%. In safety-critical scenarios, high-energy predictions can be flagged for human review.

\section{Discussion}

\textbf{Why Does Dual-Stage Compression Work?} The two stages address complementary problems. Stage 1 removes \emph{within-modality} redundancy via frequency transforms, concentrating emotion-relevant patterns. Stage 2 prevents \emph{cross-modality} shortcuts via capacity-limited query tokens, forcing genuine integration. Neither stage alone is sufficient: frequency compression without cross-modal bottleneck still allows shortcuts; bottleneck without frequency compression processes noisy, redundant inputs.

\paragraph{Theoretical Motivation for Frequency Compression.}
An information-theoretic perspective explains frequency-domain robustness. Consider random masking with rate $r$:

\emph{Time domain}: If a discriminative time step $t^*$ is masked, information loss is catastrophic---$I(x_{\text{masked}}; y) \ll I(x; y)$.

\emph{Frequency domain}: Each coefficient $X_k = \sum_t x_t e^{-2\pi ikt/T}$ encodes global temporal information. Masking distributes loss across all coefficients rather than concentrating in critical steps:
\begin{equation}
\text{Var}[I(X_{\text{masked}}; y)] < \text{Var}[I(x_{\text{masked}}; y)]
\end{equation}

This explains why wavelet/DCT preprocessing provides inherent robustness to missing data---task-relevant information is spread across frequency bands rather than concentrated in specific time steps.

\textbf{U-Shaped Multimodal Advantage.} \Cref{tab:modality} reveals a non-monotonic pattern: TAV outperforms T-only at complete data (51.6\% vs 51.0\%) and high missing rates (mr=0.9: 49.3\% vs 46.8\%), but T-only is more stable at moderate rates (mr=0.3--0.5). We hypothesize this reflects reconstruction noise: at moderate missing rates, energy-based reconstruction introduces perturbations that can disrupt text-dominant predictions; at high missing rates, the reconstruction signal becomes essential for maintaining any cross-modal information. This suggests adaptive strategies---skipping reconstruction when text confidence is high---as promising future work.

\textbf{Computational Considerations.} DCER adds moderate overhead: wavelet/DCT transforms are $O(T \log T)$, and energy iterations add $3 \times$ forward passes during inference. Total inference time is approximately 15ms vs. 12ms for MulT (25\% overhead), acceptable for most applications.

\textbf{Limitations.} Performance degrades significantly at extreme missing rates ($> 70\%$). Frequency transforms assume temporal/spatial structure and may not generalize to all modality types. Additionally, evaluation is limited to sentiment analysis; broader multimodal tasks require investigation.

\section{Conclusion}

We presented DCER, a framework unifying frequency-domain compression, cross-modal bottleneck fusion, and energy-based reconstruction for multimodal sentiment analysis. DCER achieves state-of-the-art on CMU-MOSI, CMU-MOSEI, and CH-SIMS, with intrinsic uncertainty quantification that enables reliable confidence estimation without additional calibration. Beyond sentiment analysis, the principle of exploiting modality-specific signal structure (temporal for audio, spatial for video) before cross-modal integration may benefit other multimodal tasks where such structure is known, such as audio-visual speech recognition or video question answering.

\section*{Impact Statement}

This paper presents foundational methodological advances in robust multimodal fusion and does not introduce new datasets or deploy systems on human subjects. We discuss potential broader impacts below.

\textbf{Potential positive impacts.}
Our framework improves robustness to missing modalities and provides intrinsic uncertainty quantification, which can enhance reliability in safety-critical applications such as mental health monitoring and clinical decision support. The ability to flag uncertain predictions for human review reduces the risk of automated errors in high-stakes settings.

\textbf{Potential negative impacts.}
Multimodal sentiment analysis systems process audio, visual, and textual signals that may contain sensitive personal information. Improved robustness could lower barriers to deployment in surveillance or manipulative contexts. Additionally, sentiment analysis models may inherit demographic and cultural biases present in training data, potentially producing disparate performance across populations. We encourage practitioners to conduct thorough fairness audits and obtain informed consent before deploying such systems. Our evaluation bias findings (zero-masking vs.\ noise-masking) further highlight the importance of rigorous evaluation protocols to avoid overestimating deployment readiness.

\FloatBarrier
\clearpage

\bibliographystyle{unsrt}  
\bibliography{references}

\end{document}